\documentclass[sigconf,natbib=true,anonymous=false,review=false]{acmart}
\usepackage[colorinlistoftodos]{todonotes}
\usepackage{booktabs}
\usepackage{multirow}

\AtBeginDocument{%
  }

\setcopyright{acmlicensed}
\copyrightyear{2026}
\acmYear{2026}
\acmDOI{XXXXXXX.XXXXXXX}
\acmConference[Submitted to SIGIR '26]{SIGIR}{2026}{Melbourne, Australia}
\acmISBN{978-1-4503-XXXX-X/2018/06}

\begin{document}

\title{AmharicIR+Instr: A Two-Dataset Resource for Neural Retrieval and Instruction Tuning}

\author{Tilahun Yeshambel}
\email{Tilahun.Yeshambel@uog.edu.et}
\orcid{0000-0003-0599-262X}
\affiliation{%
  \institution{Computer Science Department, Addis Ababa  University}
  \city{Addis Ababa}
  \country{Ethiopia}
}

\author{Moncef Garouani}
\email{Moncef.Garouani@irit.fr}
\orcid{0000-0003-2528-441X}
\affiliation{%
  \institution{Univ. Toulouse Capitole, IRIT, UMR5505 CNRS}
  \city{Toulouse}
  \country{France}
}

\author{Josiane Mothe}
\email{Josiane.Mothe@irit.fr}
\orcid{0000-0001-9273-2193}
\affiliation{%
\institution{UT2J, Univ. de Toulouse, IRIT, UMR5505 CNRS }
  \city{Toulouse}
  \country{France}
}

\renewcommand{\shortauthors}{Yeshambel et al.}

\newcommand{\CLS}{\texttt{[CLS]}}

\begin{abstract}
Neural retrieval and GPT-style generative models rely on  large, high-quality supervised data, which is still scarce for low-resource languages such as Amharic. We release an Amharic data resource consisting of two datasets that supports research on (i) neural retrieval-ranking and (ii) instruction-following text generation. The retrieval--ranking dataset contains 1,091 manually verified query-positive-negative document triplets  drawn from diverse Amharic sources and constructed to support contrastive training and benchmarking of neural retrievers (e.g., DPR/ColBERT-style late interaction and SPLADE-style sparse neural retrieval). Triplets are created through a combination of expert-curated queries, web-derived queries, and LLM-assisted generation, with positive/negative documents selected from the web or synthesized by LLMs and then validated by native speakers. The instruction prompt–response dataset comprises 6,285 Amharic prompt–response pairs spanning multiple domains and instruction types, generated with several LLMs and refined through manual review and correction for grammaticality, relevance, fluency, and factual plausibility. We release both datasets with standardized splits and formats (CSV/JSON/JSONL) to enable reproducible work on Amharic retrieval, ranking, and generative modelling. These datasets also come with a methodology that can be generalized to other low-resource languages. 
\end{abstract}




\received{10 February 2026}

\maketitle

\section{Introduction}

Information retrieval (IR) aims to return documents that satisfy a user’s information need. Traditional ranking functions such as TF–IDF and BM25 rely primarily on lexical matching, which can miss relevant content when queries and documents use synonyms and semantic variations, even after morphological normalization like stemming. Over the last few years, neural IR models have addressed this limitation by learning representations that capture semantic similarity and by optimizing retrieval models directly from relevance supervision. 

Neural retrieval models encode queries and documents into vectors and scores them with a similarity function. Dense dual-encoder models (DPR-style) learn a single embedding per query and document, enabling efficient approximate nearest-neighbor search \cite{karpukhin2020-dense}. Late-interaction models such as the ColBERT family retain token-level representations and compute finer-grained matching scores at retrieval time, often improving effectiveness at higher storage/compute cost \cite{Khattab2020}. Sparse neural retrievers such as the SPLADE family produce sparse, vocabulary-aligned representations through learned term expansion, combining neural modelling with inverted-index retrieval \cite{formal2021spladev2sparselexical,lassance2023}. Despite their architectural differences, these methods typically rely on supervised contrastive signals—e.g., query–positive–negative triplets or graded relevance judgments—for training and reproducible benchmarking. 
These models are effective, however, their performance depends critically on large, high-quality datasets, and models often fail to generalize, especially in low-resource languages \cite{zhang2023toward,zhao2024dense}.

Large-scale retrieval datasets such as MS MARCO~\cite{bajaj2016ms}, BEIR \cite{thakur2021beir}, and Natural Questions \cite{kwiatkowski2019} provide millions of labelled query–-document pairs or triplets, enabling both training and standardized evaluation. More recent multilingual adaptations extend retrieval research beyond English, but they often rely on translation or weak supervision  due to limited annotations and may not reflect the linguistic and domain characteristics of truly low-resource languages \cite{litschko2022,hu2023-language,goworek2025}.

Amharic, spoken by over 80 million people, is a morphologically rich Semitic language written in the 
Gez script, with substantial surface-form variation due to inflection and derivation \cite{eberhard2023}.
From an IR perspective, this variation increases vocabulary mismatch and makes tokenization and normalization decisions (e.g., segmentation, handling orthographic variants, punctuation and diacritics) particularly consequential for both sparse and neural retrievers \cite{yeshambel2020amharic,Gashaw2022,yohannes2025}. In addition, publicly available preprocessing tools and curated benchmarks are limited compared to high-resource languages, which complicates reproducibility and slows progress in neural retrieval and ranking for Amharic \cite{tonja2023}. While recent Amharic IR work has explored dataset creation and evaluation, existing resources remain limited in scale and structure, and a manually verified triplet dataset designed for contrastive training and benchmarking is still lacking \cite{yeshambel2025dense,mekonnen2025}.

Beyond retrieval, recent advances in large language models (LLMs) have broadened information access to include generative question answering, instruction following, and conversational assistants. These models are commonly trained and evaluated using  prompt–response datasets (or pairs), which are also scarce for Amharic, limiting progress in instruction tuning, generative evaluation, and retrieval-augmented generation in this language \cite{alebachew2025}.  This motivates resources that jointly support retrieval/ranking and instruction-following generation under a shared Amharic setting. In this paper, we thus release an Amharic data resource consisting of two datasets that supports research on (i) neural retrieval–ranking and (ii) instruction-following text generation. 

Here, we release the first Amharic data resource to combine (a) manually verified triplets at this scale for retrievers and (b) GPT-style prompt–response data with quality control (QC) for generative models. These datasets support reproducible IR and generation research for Amharic. The retrieval dataset contains 1,091 manually verified query–positive–negative triplets compiled from diverse Amharic sources, with accompanying documentation of query construction, relevance verification, and negative selection criteria to enable consistent contrastive training and evaluation. It is described in Section~\ref{sec:triplets}. The instruction dataset contains 6,285 Amharic prompt–response pairs covering multiple domains and instruction types; pairs were initially produced with multiple LLMs and then manually reviewed and corrected for grammaticality, relevance, fluency, and factual plausibility. It is described in Section~\ref{sec:gpt}.

We distribute both datasets with standardized formats\\ (CSV/JSON/JSONL), predefined splits, and clear dataset statements (collection process, quality control, intended uses, and limitations), so that future work can compare preprocessing and modelling choices for Amharic under a shared, well-documented experimental setup. Beyond the datasets, we also present the methodology we instantiated for Amharic and that can be generalized to other low-resource languages (Section~\ref{sec:methodology}).
Our main contributions are summarized as follows:

\begin{itemize}
    \item 	We present a reusable, native-speaker–validated data construction pipeline for low-resource languages that combines (i) expert query/prompt authoring, (ii) web-derived harvesting, and (iii) LLM-assisted generation, together with explicit positive/negative selection criteria, quality control, and normalization; we instantiate this pipeline for Amharic for both retrieval triplets and GPT-style prompt–response data.
    \item 	We release a manually verified Amharic retrieval/ranking dataset -1,091 query–positive–negative triplets- designed for neural retrievers contrastive training and benchmarking.
    \item We release an Amharic instruction-following dataset of 6,285 prompt–response pairs, with manual quality control to support instruction tuning and evaluation of generative models.
    \item 	We standardize data formats and evaluation splits \\(CSV/JSON/JSONL) and provide dataset statements and documentation, enabling reproducible comparison of retrieval, ranking, and instruction-tuned generation for Amharic.
\end{itemize}

\section{Related Work}
\label{sec:related}

MS MARCO is a large-scale machine reading comprehension and information retrieval dataset built from real user queries sampled from Bing search logs~\cite{bajaj2016ms}. It contains about one million anonymized queries and about nine million passages extracted from more than 3.5 million  retrieved documents. For each query, Bing retrieves candidate documents, from which a small set of passages (typically around ten) is automatically selected. Human editors then write natural-language answers grounded  strictly in these passages and annotate which passages support the answer using a \textit{is\_selected} tag. In training settings, passages marked as \textit{is\_selected} are treated as positive examples, while the remaining retrieved passages act as implicit negatives, reflecting realistic retrieval noise. 

Building on the large-scale retrieval setting popularized by MS MARCO, \citet{karpukhin2020-dense} show that dense retrievers can be trained at scale using weak supervision from existing QA datasets over Wikipedia. 
They construct training triples for dense retrieval by pairing each question with a positive passage and sampling negatives. 
The document collection is derived from English Wikipedia, which is preprocessed and split into fixed-length passages of 100 words, yielding approximately 21 million passages. Training queries are taken from multiple QA datasets, including Natural Questions, TriviaQA, WebQuestions, CuratedTREC, and SQuAD, comprising tens of thousands of queries. For each question, a positive passage is identified either by matching the gold Wikipedia context (when available) or by selecting the highest-ranked BM25 passage that contains the answer string. Negative passages are sampled from the remaining corpus using a combination of random negatives, BM25 hard negatives (high lexical overlap but no answer), and in-batch negatives, where positive passages of other queries in the same mini-batch serve as negatives. This weakly supervised construction enables large-scale training of dense retrievers without manual relevance annotation, and has become a standard approach for building dense retrieval datasets.

ANCE  is a dense retrieval training approach that focuses on how negative documents are mined during contrastive learning~\cite{xiong2020approximate}. Using existing datasets such as MS MARCO, Natural Questions, and TriviaQA, ANCE treats relevance judgments as positive and constructs hard negatives globally from the entire corpus. Negatives are retrieved using an approximate nearest neighbor (ANN) index over dense document embeddings, retrieving top-ranked but non-relevant documents for each query. The ANN index is asynchronously refreshed using periodically updated model checkpoints, ensuring that negatives are challenging throughout training. 

While ANCE highlights the importance of mining globally hard negatives to better match real retrieval errors, this issue becomes even more central in weakly supervised settings where explicit negatives are not provided—such as recent Amharic passage retrieval benchmark~\cite{mekonnen2025optimized}. This benchmark repurposes an existing text classification corpus \cite{azime2021amharic}. The initial text classification dataset contains more than 50k Amharic news articles spanning six domains. To simulate realistic retrieval scenarios, news headlines are treated as queries and the corresponding article bodies as passages, yielding approximately 45K query–passage pairs after preprocessing and  formatted in an MS MARCO–style retrieval format. Since explicit relevance judgments are unavailable, the dataset relies on heuristic supervision, assuming each headline is relevant to its associated article. Positive examples consist of headline–article pairs, while negative passages are implicitly provided through in-batch and retrieved non-matching passages during contrastive training. This weakly supervised construction 
does not include explicitly labelled query-positive–negative triplets. Negative documents are not part of the dataset itself but are generated implicitly during contrastive training. The retrieval benchmark splits the dataset into training and test sets. A separate validation set is not provided.

\citet{zhang2023toward} provide a best-practices guide for multilingual dense retrieval under different resource scenarios (model/data availability), highlighting the benefits of MS MARCO-style pre–fine-tuning, hard negatives, distillation, and hybrid sparse–dense retrieval—recommendations that motivate our focus on building reliable Amharic supervision and evaluation resources.

Such a dataset for Amharic was  first created by ~\citet{yeshambel2025dense}. The authors constructed a manually curated Amharic dense retrieval training dataset consisting of 152 queries only, each paired with one positive document and one negative document. Documents were collected from diverse sources (blogs, Wikipedia, news agencies, organizational websites, and social media). Each document is associated with a single topic, and relevance is defined by topic alignment. Because queries are non-overlapping, documents relevant to one query can serve as non-relevant (in-batch) negatives for others. This enables the use of in-batch negatives during training. The dataset is publicly released and is designed primarily for fine-tuning dense retrievers in low-resource settings. In contrast, Amharic still lacks a larger manually verified triplet resource designed for consistent contrastive training and benchmarking, as well as GPT-style prompt–response data for instruction tuning—gaps addressed here by our two datasets.

Existing Amharic resources are either weakly supervised or very small; none offers manually verified triplets at the size of ours, along with instruction pairs with quality control.

\section{Reusable Low-Resource Data Construction Methodology}
\label{sec:methodology}

In high-resource settings, large retrieval and instruction datasets are often constructed from abundant user logs, crowdsourcing at scale, or weak supervision from existing benchmarks (e.g., MS MARCO-style click logs or Wikipedia-based QA pairs). Such settings also make it easier to obtain large pools of implicit negatives or automatically mined training examples on which models can be trained. In contrast, for Amharic and many low-resource languages, such large-scale logs and annotated benchmarks are scarce, and weak supervision can be brittle due to morphology, orthographic variation, and limited coverage of web sources. As a result, dataset construction must place more emphasis on controlled coverage, explicit supervision, and human validation.

\subsection{Overview}
We propose a reusable methodology for constructing retrieval and instruction-following datasets in low-resource settings, when signals are unavailable or unreliable.
Our methodology combines complementary data sources (expert authoring, web-derived harvesting, and LLM-assisted generation) and enforces native-speaker validation to ensure that positives truly satisfy the information need and that negatives are genuinely non-relevant (including hard negatives with lexical overlap), language-specific normalization to reduce noise from spelling and punctuation variation, and de-duplication.  We believe these complementary steps are necessary to obtain dependable training and evaluation data in low-resource languages.

While we instantiate this methodology for Amharic in Sections~\ref{sec:triplets} and~\ref{sec:gpt}, it is language-agnostic and can be adapted to other low-resource languages by changing the domain/task templates, the web sources, and the validation guidelines.

\subsection{Pipeline Steps}
The  methodology is designed  for low-resource settings and consists in six steps as follows:
\begin{itemize}
    \item \textbf{Specify coverage}. Define target domains, task types, and linguistic phenomena of interest (e.g., morphology, lexical variation, and multi-word expressions).
    \item \textbf{Generate candidates}. Produce candidate instances using three complementary sources: expert authoring, web-derived harvesting, and LLM-assisted generation, to balance quality, realism, and topical breadth.
    \item  \textbf{Define explicit supervision}.  Establish acceptance criteria for the supervision signal: for retrieval, specify what constitutes a positive vs.  negative document (including hard negatives with lexical overlap); for generation, specify what constitutes an acceptable response (e.g., relevance, completeness, and style).
    \item  \textbf{Native-speaker validation}. Manually verify candidates for relevance/correctness and language quality (fluency and grammaticality), and check factual plausibility when prompts require factual content.
    \item  \textbf{Normalization and de-duplication}. 
    Apply language-specific normalization (punctuation, spelling conventions, white-space, and encoding) and remove duplicates or near-duplicates to reduce redundancy and noise.
    \item  \textbf{Release artifacts}. Provide  standardized formats, splits, and documentation for reproducible benchmarking and reuse.
\end{itemize}

\subsection{Instantiations and Adaptation to Other Languages}

\paragraph{Instantiation for retrieval triplets.} 
To build query-document triplets, Step~1 defines topical coverage and query types. Step~2 collects candidate documents from complementary sources (expert-driven search, web harvesting, and optional LLM-assisted generation). Step~3 specifies explicit criteria for positives and negatives: a positive document must satisfy the information need expressed by the query, whereas a negative document must not satisfy it; we additionally target hard negatives that share lexical overlap or topic proximity to the query. Step~4 performs native-speaker validation to confirm both relevance ($d^{+}$) and non-relevance
($d^{-}$). Finally, Steps~5--6 normalize text, remove duplicates, and release the triplets with documentation. Section~\ref{sec:triplets} is an instantiation to Amharic language.

\paragraph{Instantiation for GPT-style instruction data.} 
To build prompt--response pairs, Step~1 defines domain coverage and instruction-following behaviors (e.g., task completion, explanations, conversational responses). Step~2 generates candidate prompts and responses with multiple LLMs under domain/task templates. Step~3 replaces positive/negative criteria with response acceptance criteria (e.g., relevance to the prompt, completeness, fluency, and factual plausibility when required). Step~4 applies native-speaker review to correct hallucinations, unnatural phrasing, and inconsistencies while preserving the prompt intent. Steps~5--6 then normalize formatting and remove near-duplicates before releasing the dataset in standard formats.
Section~\ref{sec:gpt} is an instantiation to Amharic language.

\paragraph{Other languages.} 
To instantiate the pipeline for another low-resource language, one can reuse the same steps while replacing the domain/task templates, selecting appropriate web sources, and adjusting the validation guidelines to language-specific conventions.

\section{Amharic Query--Document Triplets for Dense Retrieval}
\label{sec:triplets}

We instantiate the low-resource data construction methodology of Section~\ref{sec:methodology} to build
manually verified Amharic query--document triplets $\langle q,d^{+},d^{-}\rangle$ for dense retrieval.

\subsection{Query Construction}
Queries are natural Amharic phrases intended to reflect realistic search behaviour. We collect queries
using three complementary approaches to balance linguistic quality, topical coverage, and scale:
(i) {Expert-curated queries}, (ii) {Topic-derived web queries}, and (iii) {LLM-generated queries}.

\paragraph{Expert-curated queries} are crafted by a native Amharic speaker to reflect diverse information
needs across domains. During query authoring, we explicitly consider Amharic linguistic properties,
including complex morphology, lexical variety, and multi-word expressions whose meanings are not
compositional (is distinct from individual words). This design ensures linguistic diversity and realistic search behaviour.

\paragraph{Topic-derived web queries} are crafted by an expert and native Amharic speaker to reflect realistic information needs across diverse domains. They are built from Amharic web pages across domains such as news,
blogs, government, technology, and related sources to capture natural user interests and topical variety. Concretely, we derive query candidates from the page/topic descriptors of documents and use this procedure to complement the manually created set with naturally occurring topics.

\paragraph{LLM-generated queries} are produced by prompting large language models to generate additional
queries for specified topics--domains. 
Controlled prompt templates are used that explicitly specified the language (Amharic), retrieval paradigm (dense retrieval), domain (e.g., sports, medical, agriculture, communication, entertainment, transport, etc.), and query characteristics (concise, information-seeking, and unambiguous). Prompts are provided in the shared repository. During generation, queries and their associated positive and negative documents are produced jointly and stored as query–positive–negative triplets in domain-specific CSV files. To ensure sufficient domain coverage and balance, we enforce a minimum of 50 generated queries per domain. All queries are then manually reviewed and lightly edited for grammaticality, spelling, and fluency.

\subsection{Positive and Negative Document Collection}
For each query, we collect one positive and one negative document. Unlike query construction (which we categorize by generation strategy), we organize document collection by the source from which documents are obtained:

\paragraph{Search engine--retrieved documents.} For expert--curated queries, we execute each query using Google Search and inspect the top ranked results. Positive documents are selected by carefully reading the top results and choosing the most relevant document(s). Negative documents are chosen by reading non-relevant results; while many negatives occur lower in the ranking, some appear near the top and share query terms without satisfying the information need. We tag such cases as hard negatives.

\paragraph{Website-sourced documents.} For web-derived queries (collected from web pages), we assign the corresponding source page as the positive document, ensuring that the query directly reflects the document content. Negative documents are sampled from the same website but from non-relevant pages, so that negatives preserve realistic topical and stylistic properties (i.e., realistic retrieval noise).

\paragraph{LLM-synthesized documents.}
For each LLM-generated query, the corresponding positive and negative (including hard-negative) documents are generated simultaneously using controlled LLM prompts and stored together with the query in CSV files organized by domain. The prompts explicitly constrains the output language to Amharic and the document length to 5–7 sentences. Positive documents are required to satisfy the query’s information need, whereas negative documents are topically related but semantically misaligned. Prompts are provided in the reporsitory. Following automatic generation, documents are manually inspected and edited to improve linguistic quality and consistency. 

This methodology ensures that each query type is paired with high-quality positive and negative documents, supporting robust training and evaluation of dense retrieval  in the Amharic language. 

\subsection{Triplet Construction and Validation}
\label{sec:triplet_construction}
Each instance is a triplet $\langle q, d^{+}, d^{-} \rangle$, which supports contrastive learning for dense retrieval models. A document is labeled positive if it provides direct semantic evidence that addresses the query; otherwise it is labeled negative. Importantly, we do not rely solely on search engine rank or LLM generation labels: all positives and negatives are manually assessed. Annotators carefully read each candidate document with respect to its query and verify (i) that $d^{+}$ satisfies the query and (ii) that $d^{-}$ does not satisfy the query, including in the presence of lexical overlap (hard negatives).

\begin{table}[t]
\centering
\caption{Dense retrieval triplet dataset statistics.}
\label{tab:triplet_stats}
\begin{tabular}{lr}
\toprule
Statistic & Value \\
\midrule
\#queries & 1,091 \\
\#documents & 2,182 \\
\#sentences & 43,055 \\
\#words & 634,963 \\
Avg.\ query length (words) & 6 \\
Avg.\ document length (words) & 291 \\
Vocabulary size (unique words) & 106,270 \\
Avg.\ \#sentences per document & 19 \\
Min.\ \#sentences per document & 5 \\
Max.\ \#sentences per document & 231 \\
\bottomrule
\end{tabular}
\end{table}

\begin{figure}[t]
  \centering
  \includegraphics[width=0.95\linewidth]{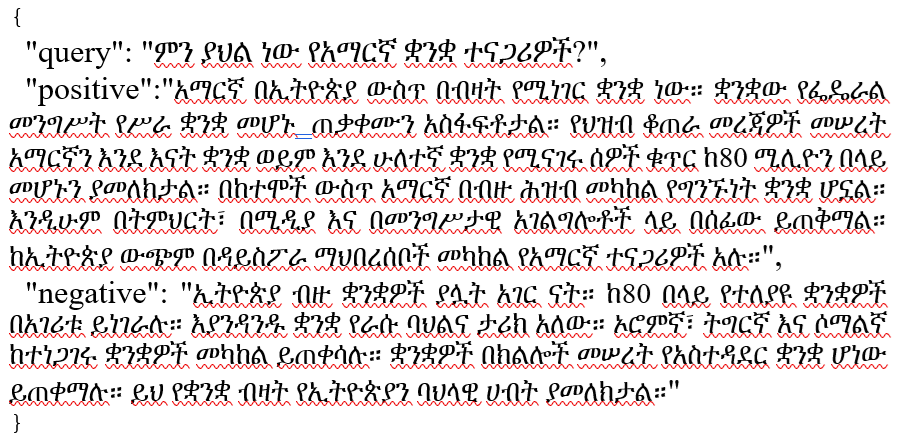}
  \caption{Example of a JSON record in the released triplet dataset.}
  \label{fig:json_example}
\end{figure}

\paragraph{Dataset format and statistics.}
In total, the dataset consists of 1,091 triplets, each composed of one query, one positive document, and one negative document.  Among these, 200 triplets correspond to manually created queries, 251 triplets are derived from web-collected queries, and 640 triplets are synthesized using large language models. 
Table~\ref{tab:triplet_stats} reports some dataset-level statistics. All text is in UTF-8 and written in Amharic (Ge\textquotesingle ez) script. We release the dataset in JSON and CSV formats, with one triplet per line. Figure~\ref{fig:json_example} is an example of a JSON record.

\paragraph{Preprocessing.}
We preprocess queries and documents to remove duplicates, incomplete entries, non-Amharic content, and noise such as HTML tags and irrelevant metadata. We normalize punctuation and white-space, map
sentence boundary markers ``::'' and ``.'' to the standard Amharic full stop, 
and collapse long whitespace/newline sequences into a single space. We also perform manual checks for language correctness and fluency and filter sensitive or personal information.

\section{Amharic GPT-Style Prompt--Response Dataset}
\label{sec:gpt}
Using the same pipeline principles (Section~\ref{sec:methodology}), we construct a GPT-style Amharic prompt--response dataset of 6,285 pairs. To maximize coverage while maintaining quality, we generate candidate pairs with multiple large language models (LLMs), including ChatGPT, Gemini, and DeepSeek, and then perform manual validation and correction by native Amharic speakers. 

\subsection{Data Construction}
\paragraph{Domains and task types.}
To encourage broad generalization, we predefine a set of domains including education, health, agriculture, governance, culture, religion, technology, and daily life.
Within each domain, we focus on common instruction-following behaviors, including (i) direct task completion (e.g., ``write/translate/summarize''), (ii) explanation (e.g., ``explain why/how''), and (iii) conversational responses (e.g., multi-turn assistant-style answers). These domain and task specifications guide prompt design and ensure consistent coverage across different LLMs.

\paragraph{Prompt and response generation.}
Prompt--response candidates are generated using LLMs under domain- and task-level guidance. Rather than manually authoring each prompt, we provide the target domain and task type and ask the LLMs to produce diverse Amharic prompts together with appropriate responses. This strategy increases topical diversity while keeping the dataset structure controlled.  

\paragraph{Manual review and quality control.}
Following Steps~4--5 of Section~\ref{sec:methodology}, native Amharic speakers review all pairs for grammaticality, fluency, and relevance to the prompt, and correct hallucinations or unnatural phrasing while preserving the prompt intent.
We then normalize formatting and remove duplicates and near-duplicates. 

\subsection{Format and Release}
\paragraph{Dataset refinement.}
After manual correction, we normalize punctuation, spelling conventions, and formatting. We remove duplicates and highly similar pairs to reduce redundancy. The final dataset consists of clean, high-quality Amharic prompt--response pairs suitable for both training and evaluation.

\paragraph{Format.}
We release the dataset in both CSV and JSON formats. The CSV format supports tabular inspection and preprocessing, while the JSON format facilitates integration with common LLM training pipelines.

\begin{table*}[t]
\centering
\caption{Retrieval performance of fine-tuned Splade, RoBERTa, and ColBERT models at different cutoff values $K$.}
\label{tab:triplet_sources_transposed}
\small
\setlength{\tabcolsep}{4pt}
\begin{tabular}{c|cccc|cccc|cccc}
\toprule
 & \multicolumn{4}{c|}{\textbf{Splade}} 
 & \multicolumn{4}{c|}{\textbf{RoBERTa}} 
 & \multicolumn{4}{c}{\textbf{ColBERT}} \\
K 
 & NDCG & MAP & Recall & Precision
 & NDCG & MAP & Recall & Precision
 & NDCG & MAP & Recall & Precision \\
\midrule
1   & .598 & .598 & .598 & .598 & \textbf{.744} & \textbf{.744} & \textbf{.744} &\textbf{ .744} & .666 & .666 & .666 & .666 \\
3   & .712 & .684 & .794 & .264 & \textbf{.817} & \textbf{.798} & \textbf{.872} & \textbf{.290} & .752 & .732 & .812 & .270 \\
5   & .735 & .696 & .849 & .169 & \textbf{.832} & \textbf{.806 }& \textbf{.908} & \textbf{.181} & .773 & .743 & .863 & .172 \\
10  & .769 & .705 & .913 & .091 & \textbf{.845} & \textbf{.812} & \textbf{.949} & \textbf{.094 }& .787 & .749 & .904 & .090 \\
100 & .772 & .708 & .977 & .009 & \textbf{.853} & \textbf{.814} & \textbf{.986} & \textbf{.009} & .803 & .752 & .981 & .009 \\
\bottomrule
\end{tabular}
\end{table*}

\section{Intended Uses and Baselines}
\label{sec:intended_uses}

Our two datasets (dense retrieval triplets and GPT-style prompt--response pairs) are intended to support research in ad-hoc information retrieval and Amharic text generation, with a focus on reproducible evaluation and model development for low-resource languages.
Both datasets will be released on Hugging Face and GitHub upon paper acceptance, together with documentation and access instructions to support reproducibility and further research on Amharic and other low-resource languages.

\subsection{Dense retrieval triplets}
The Amharic query--document triplet dataset enables training and evaluation of retrieval models that rely on explicit positive and negative examples. It can be used to:
\begin{itemize}
    \item Benchmark dense and late-interaction retrieval models under a common Amharic testbed,
    \item Train models with contrastive objectives using manually verified triplets, and
    \item Study the impact of hard negatives, lexical overlap, and morphology on retrieval.
\end{itemize}

The dataset is compatible with common retrieval paradigms, including dual-encoder dense retrievers (e.g., DPR-style bi-encoders), sentence embedding approaches (e.g., Sentence-BERT variants), hard-negative mining methods (e.g., ANCE-style training), and late-interaction models such as ColBERT/ColBERTv2. It can also be used to compare with sparse and hybrid baselines, including lexical retrieval and sparse expansion methods (e.g., BM25 and SPLADE).

\subsection{GPT-style prompt--response pairs}
The Amharic prompt--response dataset supports instruction tuning, supervised fine-tuning, and controlled evaluation of generative models in Amharic. Potential uses include:
\begin{itemize}
    \item training or adapting Amharic-capable LLMs to follow instructions across domains,
    \item evaluating generation quality (fluency, relevance, robustness) under domain shifts, and
    \item studying retrieval-augmented generation (RAG) by combining the prompt--response dataset with the retrieval triplets as an external knowledge source.
\end{itemize}

\subsection{Broader impact}
Together, these resources lower the barrier to developing and evaluating retrieval and generation systems for Amharic and provide a practical foundation for studying LLM adaptation and evaluation in low-resource settings.

\subsection{Baselines}

\paragraph{Dataset:}
We consider here our dataset which contains 1,091 queries, 2,182 Amharic documents, and relevance judgments (qrels) for all query–document triplets. The dataset is specifically designed for fine-tuning and evaluating neural retrieval models and  is split into 654 training queries, 218 validation queries, and 219 testing queries. The qrels were used to evaluate models performance using standard IR metrics, including Mean Average Precision (MAP), Precision, Recall, and NDCG.

\paragraph{Models:} We  fine-tuned the SPLADE-RoBERTa-Amharic-Medium (sparse expansion), RoBERTa-Amharic-Embed-Medium (dense embeddings), and ColBERT-Amharic-Medium (late interaction) pre-trained Amharic neural retrieval models\,\footnote{Those models are available on\,: {\url{https://huggingface.co/rasyosef/[ModelName]}}} on the training/validation part of our dataset presented in Sec.~\ref{sec:triplets}. Sparse expansion models improve lexical coverage of queries by expanding input tokens into weighted representations over the vocabulary, bridging gaps between query and document terms. Dense embedding models represent queries and documents as continuous vectors in a semantic space, allowing fine-grained similarity computation. Late interaction models combine token-level embeddings with efficient similarity aggregation for per-token matching. These models are further fine-tuned on our triplet-specific training dataset contains query–positive–negative triplets. This training allows the models learn to rank specifically according to our relevance judgments.
The fine-tuned models and the dataset splits will be available on GitHub.

\paragraph{Results:}Table~\ref{tab:triplet_sources_transposed} reports the results on the test part of our dataset. All models achieve competitive retrieval performance across most cut-off values. RoBERTa achieves the highest performances on the different cutoff values. ColBERT performed slightly lower but maintains strong Recall, and SPLADE is the less performing. NDCG, MAP and recall improve with increasing K, while precision decreased.  The important decrease of precision with cut-off $K>3$ is because each query has only a small number of relevant documents in qrels, causing more non-relevant documents to appear among the top-ranked results. 
These results establish a strong baseline for future experiments.

\section{Conclusion}
Neural retrieval and GPT-style generative models rely on large, high-quality supervised datasets, which are scarce for low-resource languages like Amharic. 
We present a new Amharic resource consisting of two datasets for (i) neural retrieval and ranking, and (ii) instruction-following text generation.  The retrieval–ranking dataset contains query–-positive-–negative triplets for contrastive training and benchmarking of neural retrievers, including ColBERT-style late interaction and SPLADE-style sparse models. 
Triplets are created using expert-curated queries, web-derived queries, and LLM-assisted generation, with documents validated by native speakers. The instruction prompt–response dataset includes Amharic prompt–response pairs across multiple domains, generated with three LLMs, and refined for grammaticality, relevance, fluency, and factual accuracy. Baseline experiments on fine-tuned SPLADE, RoBERTa, and ColBERT models show that RoBERTa achieves the best retrieval performance. The results indicate that fine-tuning on the Amharic triplet dataset yields effective retrieval performance across multiple neural architectures. Both datasets are released with standardized splits and formats (CSV/JSON/JSONL) to enable reproducible research. The methodology can be generalized to other low-resource languages, helping expand NLP resources beyond widely studied languages. 

These datasets and methods address the scarcity of high-quality Amharic data and provide a foundation for future work in neural retrieval and generative modeling.

\bibliographystyle{ACM-Reference-Format}
\bibliography{AmharicResource}

\end{document}